\documentclass[journal]{IEEEtran}

\ifCLASSINFOpdf
\else
   \usepackage[dvips]{graphicx}
\fi
\usepackage{url}

\usepackage{epsfig}
\usepackage{subcaption}
\usepackage{multirow}
\usepackage{amssymb}
\usepackage{graphicx}
\usepackage{makecell}
\usepackage{comment}

\usepackage[pagebackref=true,breaklinks=true,colorlinks,bookmarks=false]{hyperref}

\makeatletter
\DeclareRobustCommand\onedot{\futurelet\@let@token\@onedot}
\def\@onedot{\ifx\@let@token.\else.\null\fi\xspace}

\makeatother

\begin{document}

\title{Towards NIR-VIS Masked Face Recognition}
\author{Hang Du,~\IEEEmembership{}
        Hailin Shi,~\IEEEmembership{Member, IEEE,}
        Yinglu Liu,~\IEEEmembership{Member, IEEE,}
        Dan Zeng,~\IEEEmembership{Member, IEEE,} \par
        and Tao Mei~\IEEEmembership{Fellow, IEEE} 
\thanks{H. Du and D. Zeng are with the Key Laboratory of Specialty Fiber Optics and Optical Access Networks, Joint International Research Laboratory of Specialty Fiber Optics and Advanced Communication, Shanghai Institute of Advanced Communication and Data Science, Shanghai University, Shanghai, 200444, China (e-mail: duhang@shu.edu.cn, dzeng@shu.edu.cn)}
\thanks{H. Shi, Y. Liu and T. Mei are with the JD AI Research, Beijing, 100029, China (e-mail: shihailin@jd.com, liuyinglu1@jd.com, tmei@live.com)}
\thanks{Corresponding author: Dan Zeng. }

}

\maketitle

\begin{abstract} 
Near-infrared to visible (NIR-VIS) face recognition is the most common case in heterogeneous face recognition, which aims to match a pair of face images captured from two different modalities. 
Existing deep learning based methods have made remarkable progress in NIR-VIS face recognition, while it encounters certain newly-emerged difficulties during the pandemic of COVID-19, since people are supposed to wear facial masks to cut off the spread of the virus. 
We define this task as NIR-VIS masked face recognition, and find it problematic with the masked face in the NIR probe image.
First, the lack of masked face data is a challenging issue for the network training. Second, most of the facial parts (cheeks, mouth, nose~\textit{etc.}) are fully occluded by the mask, which leads to a large amount of loss of information. Third, the domain gap still exists in the remaining facial parts.
In such scenario, the existing methods suffer from significant performance degradation caused by the above issues. 
In this paper, we aim to address the challenge of NIR-VIS masked face recognition from the perspectives of training data and training method.
Specifically, we propose a novel heterogeneous training method to maximize the mutual information shared by the face representation of two domains with the help of semi-siamese networks. In addition, a 3D face reconstruction based approach is employed to synthesize masked face from the existing NIR image. 
Resorting to these practices, our solution provides the domain-invariant face representation which is also robust to the mask occlusion.
Extensive experiments on three NIR-VIS face datasets demonstrate the effectiveness and cross-dataset-generalization capacity of our method. 
\end{abstract}


\IEEEpeerreviewmaketitle

\section{Introduction}

\begin{figure}
    \centering
    \includegraphics[scale=0.38]{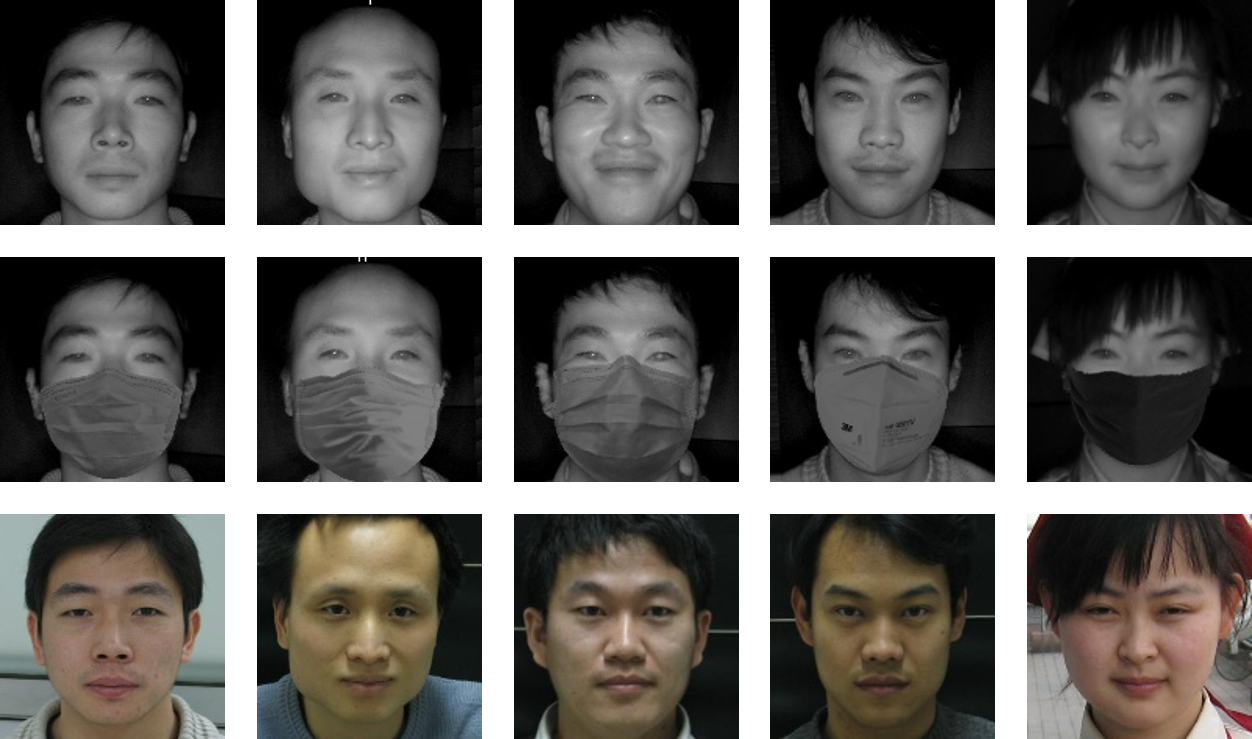}
    \caption{Examples of NIR faces, synthesized masked NIR faces and VIS faces (from top to bottom). The facial appearance variation is enlarged by the facial mask occlusion. }
    \label{figure}
    \vspace{-2em}
\end{figure}

\IEEEPARstart{N}{ear-infrared} to visible (NIR-VIS) face recognition has been widely adopted in many face recognition applications, especially on the condition of low illumination. 
It aims to match a near-infrared (NIR) probe face image with a visible (VIS) gallery face image. 
Existing deep learning based methods~\cite{Liu2016TransferringDR,he2017learning,lezama2017not,yu2019pose,duan2020cross} have made remarkable progress in NIR-VIS face recognition.
However, at the pandemic of novel coronavirus 2019 (COVID-19), humans are supposed to wear facial masks to cut off the spread of the virus. Hence, a masked NIR probe face is required to be matched with a VIS gallery face. We define this task as NIR-VIS masked face recognition, and find it problematic with the masked face in the NIR probe image. 
First, the lack of masked face data is a challenging issue for the existing training methods. 
Second, as shown in Fig.~\ref{figure}, since most of the facial parts are fully occluded by the mask, a large amount of information is lost in such case. 
Third, we can also observe that the domain gap still exists in the remaining facial parts. 
The above issues result in the significant performance degradation on NIR-VIS masked face recognition.
Therefore, it is urgent to address these newly-emerged difficulties in NIR-VIS masked face recognition. 

Recently, many deep learning methods have been proposed for NIR-VIS face recognition, which can be divided into two schemes. 
The first scheme focuses on learning face representation from the heterogeneous data.  
They~\cite{Saxena2016HeterogeneousFR,Liu2016TransferringDR,reale2016seeing} aim to learn a common feature representation space in which the face representation of the same identity from two domains is similar. The typical routine is to pretrain CNN on the large-scale visible images and fine-tune it on the heterogeneous data. 
Besides, learning domain-invariant representation is another choice. IDR~\cite{he2017learning} and W-CNN~\cite{he2018wasserstein} achieve this by dividing the high-level convolutional layers into two orthogonal subspaces. 
However, the above methods suffer from two issues in NIR-VIS masked face recognition. On one hand, the domain gap bewteen NIR and VIS images is enlarged due to the occlusion of mask, which makes it difficult to effectively learn face representation. On the other hand, the limited training data will lead to the overfitting problem. Therefore, these methods can not perform well on NIR-VIS masked face recognition.

The second scheme aims to synthesize face images from one domain to another for reducing the domain gap.  
They~\cite{juefei2015nir,lezama2017not,song2017adversarial,Di2018PolarimetricTT,yu2019pose,Di2019PolarimetricTT,duan2020cross,Di2021MultiScaleTT} propose to synthesize the VIS images from the NIR or thermal images, and then perform the regular face recognition algorithm in the VIS domain.
With the significant improvement of image generation, the above methods have achieve state-of-the-art performance in general NIR-VIS face recognition. 
However, in the scenario of NIR-VIS masked face recognition, they fail to synthesize the photo-realistic full VIS faces from the masked NIR faces, since most of the facial information is lost due to the mask occlusion. 

In addition, with the pandemic of COVID-19, certain methods~\cite{geng2020masked,ding2020masked} have been proposed for general masked face recognition. Geng~\textit{et al.}~\cite{geng2020masked} introduce a GAN-based method to synthesize masked face and a domain constrained loss to make the masked faces close to its corresponding full face in the feature space. 
Besides, a latent part detection model~\cite{ding2020masked} is proposed to locate the facial region which is robust to mask wearing and used to extract discriminate features. 
The above methods mainly study the general homogeneous masked face recognition, but the large domain gap between masked NIR faces and full VIR faces is a more challenging task. 

In this paper, we study the NIR-VIS masked face recognition which is a challenging task needed to be addressed at the pandemic of COVID-19. We intend to address it from the perspectives of training data and training method. 
First, a heterogeneous semi-siamese training (HSST) method is proposed to maximize the mutual information shared by the face representation of two domains with the help of semi-siamese networks~\cite{du2020semi}. Specifically, a positive pair of NIR-VIS face images are fed into the semi-siamese networks; with the optimization of the learning objective, the semi-siamese networks enable to maximize the mutual information shared by the face representation of masked NIR images and VIS images. 
Since two heterogeneous prototypes are employed to compute the training loss, they can provide two complementary views to maximize the mutual information. 
Second, to obtain realistic masked face data, we adopt a 3D face reconstruction based approach to synthesize masked face from the existing  images. 
Resorting to the above practices, our solution provides the domain-invariant face representation which is also robust to the mask occlusion.
Extensive experiments on CASIA NIR-VIS 2.0, Oulu-CASIA NIR-VIS, and BUAA-VisNir datasets demonstrate the effectiveness and cross-dataset-generalization capacity of our method.

\section{Method} 

\subsection{Heterogeneous Semi-Siamese Training} 
Semi-siamese training (SST)~\cite{du2020semi} is proposed to handle shallow face learning. Since the extreme lack of intra-class diversity, traditional training methods suffer from the model degeneration and overfitting issue on shallow face learning. To address these problems, semi-siamese training consist of a probe-net $\phi_{p}$ to embed the features of probe images and a gallery-net $\phi_{g}$ to update prototypes queue by the features of gallery images. The probe features and the feature-based prototype queue are used to compute the training losses. The probe-net is optimized by the SGD and the gallery-net is updated by the moving-average. 

\begin{figure}
    \centering
    \includegraphics[scale=0.18]{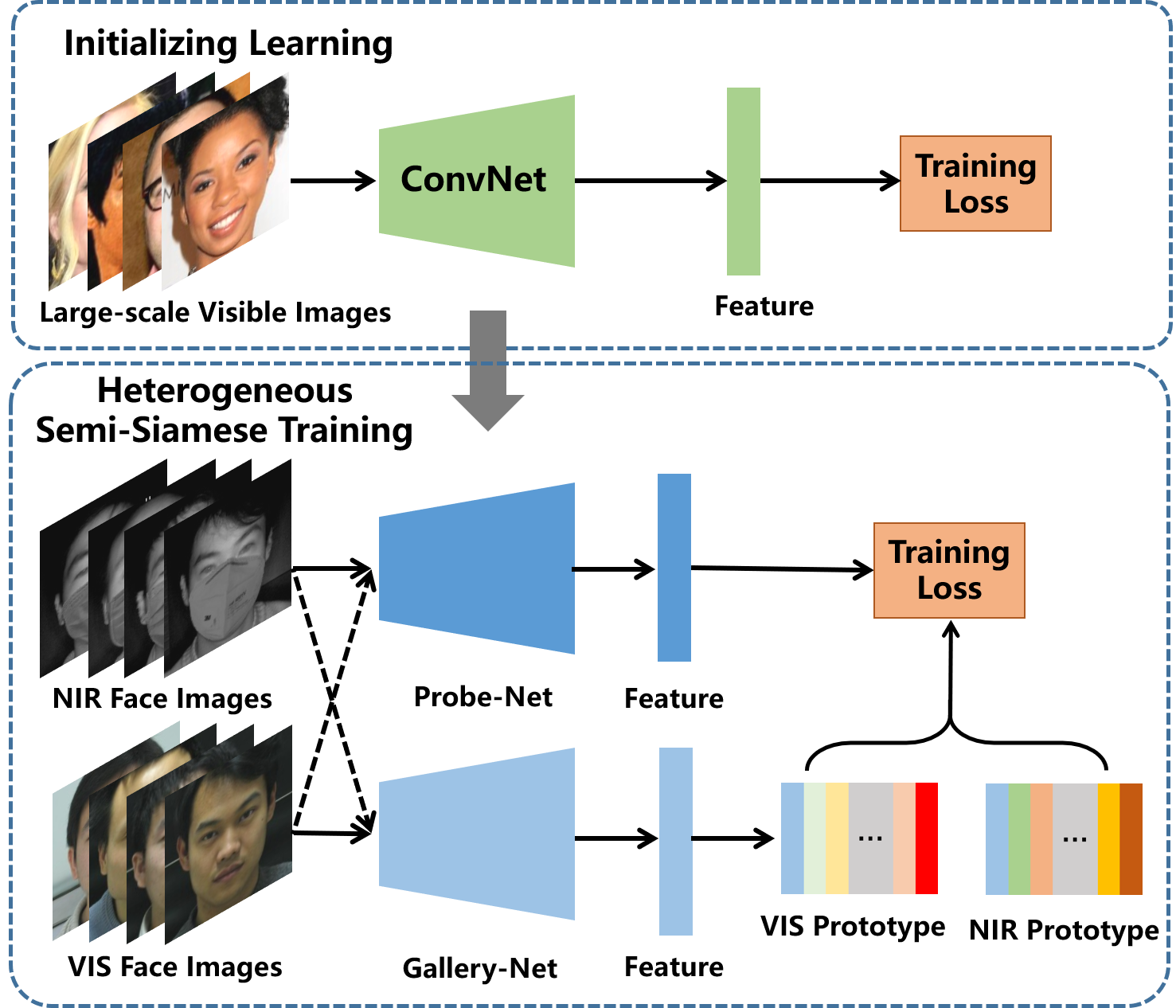}
    \caption{Overview of our training method. We first pre-train the model on the large-scale visible face images, and then fine-tune it by heterogeneous semi-siamese training. The solid line arrows and crossing dash arrows refer to a positive pair of NIR and VIS images have an even chance to be fed into the probe-net or the gallery-net respectively.
    }
    \label{method}
    \vspace{-1.5em}
\end{figure}

Different to the shallow face learning, the intra-class diversity in NIR-VIS masked face recognition is large due to the domain gap and the loss of facial information. 
Based on the semi-siamese networks, we propose a heterogeneous semi-siamese training (HSST) for NIR-VIS masked face recognition. Compared to the orginal SST, we feed a positive pair of heterogeneous faces, including a masked NIR face and a full VIS face, into the semi-siamese networks respectively (as shown in Fig.~\ref{method}). 
Besides, we construct two heterogeneous prototype queues that contain the features of NIR face and VIS face inferred by the gallery-net, respectively. 
Then, the NIR prototype queue is employed to compute the training loss with the feature of VIS images inferred by the probe-net, and the VIS prototype queue is used to compute the training loss with the feature of NIR images. 
The two networks are pretrained on the visible face dataset.

In the following, we study how HSST can perform well on NIR-VIS masked face recognition. Generally, the classification scheme is a typical routine for deep face representation learning.  
We take the softmax cross-entropy loss function (omitting the bias term) as an example, which can be formulated as: 
\begin{equation}
\mathcal{L}=- \log \frac{e^{s \cos \left(\theta_{i, y}\right)}}{e^{s \cos \left(\theta_{i, y}\right)}+\sum_{j=1, j \neq y}^{n} e^{s \cos \left(\theta_{i, j}\right)}},
\end{equation} 
where $\cos(\theta_{i, y})$ is the cosine similarity between feature $x_i$ and its ground-truth prototype $w_y$, $s$ is the scale factor, and $n$ is the number of prototypes. 

Let $I_{N}$ and $I_{V}$ be a positive pair of NIR and VIS face images respectively. Suppose that we input $I_{N}$ into the probe-net and $I_{V}$ into the gallery-net. In this condition, the softmax loss can be reformulated by:
\begin{equation}
\mathcal{L}(I_{N}, I_{V})= - \log \frac{e^{s \phi_{p}(I_{N}) \phi_{g}(I_{V})}}
{ e^{s \phi_{p}(I_{N}) \phi_{g}(I_{V})} +
\sum_{j=1}^{n} e^{s \phi_{p}(I_{N}) f^{V}_{j}}},
\end{equation}
where $\phi_{p}(I_{N})$ is the NIR face representation inferred by the probe-net, $\phi_{g}(I_{V})$ is the VIS face representation inferred by the gallery-net, and $f^{V}_{j}$ is the $j$th feature of the VIS prototype queue. 
For each iteration, the sampled training ID is disjoint to the ID in prototype queue. Thus, there are one positive pair and $n$ negative pairs in the above loss function. Besides, the $\phi_{g}(I_{V})$ is not the same as $f^{V}_{j}$, since they are inferred by the gallery-net of different states.
Then, we denote the mutual information bewteen the NIR and VIS face representations as $\mathcal{I}(\phi_{p}(I_{N}) ; \phi_{g}(I_{V}))$. 
Minimizing the learning objective $\mathcal{L}(I_{N}, I_{V})$ is equivalent to maximize the mutual information which can be formulated as: 
\begin{equation}
\mathcal{I}(\phi_{p}(I_{N}) ; \phi_{g}(I_{V})) = 
\mathcal{H}(\phi_{p}(I_{N})) - \mathcal{H}(\phi_{p}(I_{N}) | \phi_{g}(I_{V})) ,
\end{equation} 
where $\mathcal{H}(\cdot)$ denotes the entropy. By minimizing the training loss $\mathcal{L}(I_{N}, I_{V})$, the face representation is spread in the feature space, and those of the same identity from two domains is becoming close. 
In other words, it implicitly maximizes the entropy $\mathcal{H}(\phi_{p}(I_{N}))$ and minimizes the conditional entropy $\mathcal{H}(\phi_{p}(I_{N}) | \phi_{g}(I_{V}))$ during the training procedure. 

Whereas NIR face and VIS face have an even chance to be fed into the probe-net or the gallery-net, the heterogeneous prototype queues are utilized to compute two training losses,~\textit{i.e.,} $\mathcal{L}(I_{N}, I_{V})$ and $\mathcal{L}(I_{V}, I_{N})$, which are minimized simultaneously during the training process. They provide two complementary views to maximize the mutual information of the NIR and VIS face representations. 
In this way, HSST fulfils maximizing both $\mathcal{I}(\phi_{p}(I_{N}) ; \phi_{g}(I_{V}))$ and $\mathcal{I}(\phi_{p}(I_{V}) ; \phi_{g}(I_{N}))$, which facilitates the semi-siamese networks to improve the quality of domain-invariant face representation that is also robust to the mask occlusion. 
Besides, it is worth noting that we only take the probe-net for the tests, which guarantees the same inference efficiency and comparison fairness with the single network design. 

\subsection{3D Face Reconstruction based Masked Face Synthesis} 
\begin{figure}
    \centering
    \includegraphics[scale=0.3]{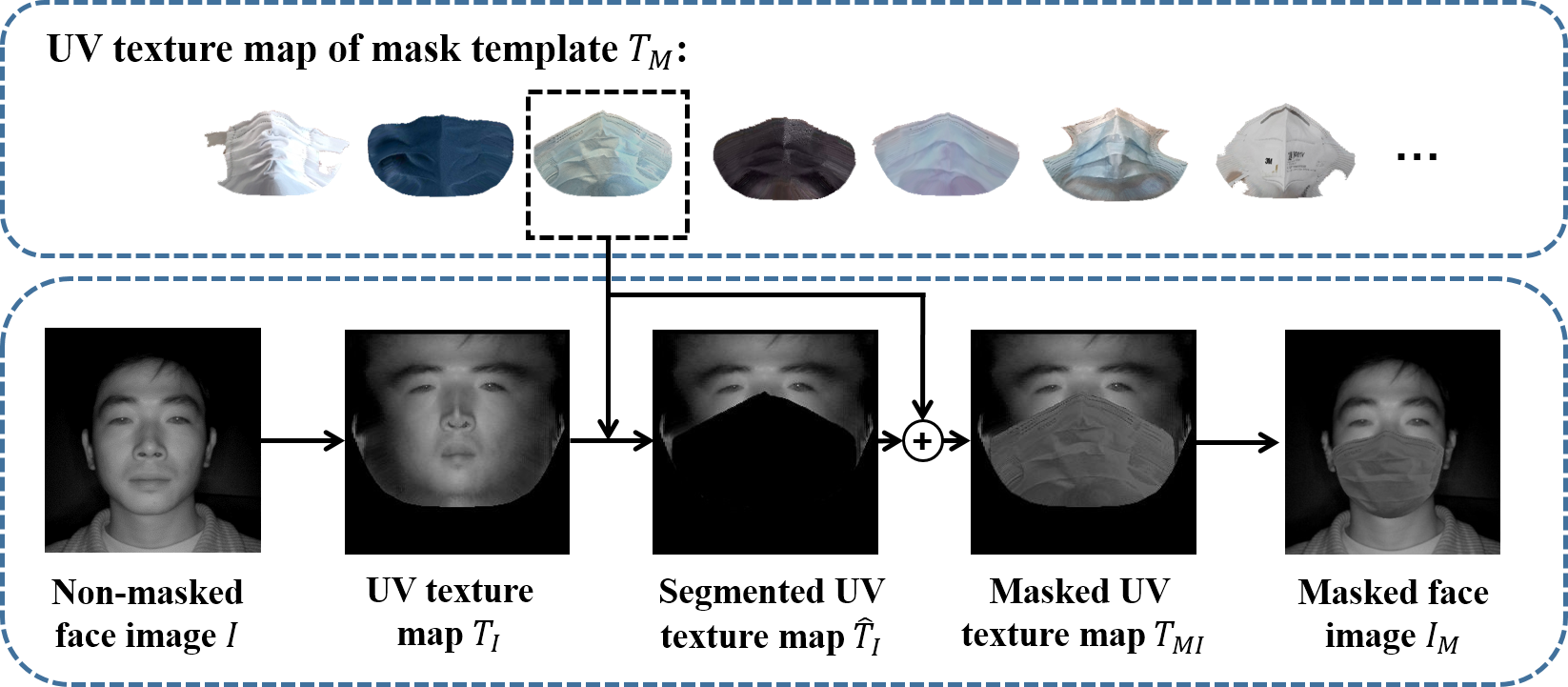}
    \caption{The pipeline of masked face synthesis. First, we obtain the different kinds of mask templates. Then, we conduct 3D face reconstruction and add the mask template on the UV texture map of the non-masked face. Finally, we recover the 2D masked face image from the masked UV texture map.  }
    \label{3d_masked}
    \vspace{-1.5em}
\end{figure}
Since collecting the realistic masked faces is expensive, we present a 3D face reconstruction based method to synthesize masked face. Our method employs PR-Net~\cite{feng2018joint} to extract the UV texture map and its corresponding UV position map to represent the 3D face. 
Fig.~\ref{3d_masked} shows the pipeline of our method to synthesize the masked face. Specifically, we first segment the facial mask from the real masked face images and obtain the UV texture map of mask template $T_{M}$. Then, given a non-masked face image $I$, we obtain its UV texture map $T_{I}$ and UV position map $P_{I}$, and remove the corresponding region on the UV texture map $T_{I}$ according to the mask template, and obtain the remaining UV texture map $\hat T_{I}$. Finally, we add the mask template $T_{M}$ on $\hat T_{I}$. This operation can be simply formulated as:
\begin{equation}
 T_{MI} = T_{M} + \hat T_{I},
\end{equation} 
where $T_{MI}$ is the UV texture map of masked face images. Then, we recover the 2D masked face images $I_{M}$ from the UV texture map $T_{MI}$ and the UV position map $P_{I}$. 

Compared with the 2D-landmark-based and GAN-based masked face generation methods~\cite{geng2020masked,ding2020masked}, we consider 3D face reconstruction is a more accurate practice for masked face synthesis, especially for the large-pose case. 

\section{Experiments} 

\begin{table}[tp]
    \begin{center}
        \caption{Performance (\%) comparison on the 1-fold of the non-masked and synthesized masked CASIA NIR-VIS 2.0. }
    \label{org_vs_masked}
    \resizebox{0.9\linewidth}{!}{
    \begin{tabular}{|c|c||c|c|}
    
    \hline
         \makecell[c] {Training Data} &
         \makecell[c] {Test  Data }&
         \makecell[c] {Rank-1 \\ Accuracy}& 
         \makecell[c] {VR \\ @FAR=0.1\% }\\
    \hline\hline
    \multicolumn{4}{|c|}{Plain Training}\\
    \hline
       Non-masked & Non-masked & 98.25 & 97.74\\
    \hline   
        Non-masked & Masked & 53.88 & 42.36\\        
    \hline   
       Masked & Masked &  83.90 & 77.23\\
    \hline \hline
    \multicolumn{4}{|c|}{HSST}\\
    \hline 
        Non-masked & Non-masked & 99.30 & 98.79\\
    \hline   
        Non-masked & Masked & 67.52 & 59.35\\ 
    \hline   
        Masked & Masked& 93.60 & 90.25\\
    \hline     
    \end{tabular}}
    \end{center}
    \vspace{-2em}
\end{table}

\subsection{Datasets and Preprocessing}
To demonstrate the effectiveness of our method, we employ three widely-used NIR-VIS face datasets, including the CASIA NIR-VIS 2.0~\cite{li2013casia}, the Oulu-CASIA NIR-VIS~\cite{chen2009learning}, and the BUAA-VisNir~\cite{huang2012buaa} datasets. 
Among them, CASIA NIR-VIS 2.0 dataset is the largest NIR-VIS face dataset, including 725 identities with 17,580 face images. 
Oulu-CASIA NIR-VIS dataset consists of 80 identities with 6 different expression, and each identity contains 48 NIR images and 48 VIS images. The training set and test set respectively contain 20 identities. 
BUAA-VisNir dataset contains 150 identities with 9 NIR images and 9 VIS images per identity. The training set contains 50 identities with 900 images and the test set contains the reminder of 1800 images. 
Since the NIR face images are used as the probe images in the test protocol, we add the masks on all the NIR images in the training and test sets. Besides, we utilize MS1M-v1c\footnote{A cleaned version of MS-Celeb-1M: http://trillionpairs.deepglint.com}~\cite{guo2016ms}
as the pre-training dataset. All the faces are detected by Faceboxes~\cite{zhang2017faceboxes}. Then, we align and crop them to 144$\times$144 image according to the five facial points. 

\begin{table*}[h]
    \begin{center}
        \caption{Performance (\%) comparison on the 1-fold of the 
         synthesized masked CASIA NIR-VIS 2.0 dataset, the synthesized masked Oulu-CASIA NIR-VIS dataset, and the synthesized masked BUAA-VisNir dataset. }
    \label{cross_dataset}
    \resizebox{0.9\textwidth}{!}{
    \begin{tabular}{|c||c|c|c|c|c|c|c|}
    \hline
        \multirow{3}{*}{Method} & 
        \multicolumn{2}{c|}{CASIA NIR-VIS 2.0}&
        \multicolumn{2}{c|}{Oulu-CASIA NIR-VIS}&
        \multicolumn{3}{c|}{BUAA-VisNir}
        \\ 
        \cline{2-8}&\makecell[c] {Rank-1 \\ Accuracy}& \makecell[c] {VR\\ @FAR=0.1\% }&\makecell[c] { VR \\ @FAR=1\% }& \makecell[c] {VR\\ @FAR=0.1\% }&\makecell[c] {Rank-1 \\ Accuracy}& \makecell[c] {VR\\ @FAR=1\% }&\makecell[c] {VR\\ @FAR=0.1\% }\\
    \hline\hline
        softmax & 93.81 & 90.83 & 84.4 & 61.6 & 92.6 & 62.3 & 43.0 \\
    \hline   
        AM-softmax & 95.00 & 92.38 & 85.4 & 63.8 & 93.0 & 67.8 & 46.5 \\
    \hline   
        Arc-softmax & 94.63 & 91.73 & 85.7 & 64.4 & 93.3 & 68.4 & 46.7 \\
    \hline 
        Triplet & 97.40 & 96.43 & 86.9 & 65.0 & 97.2 & 78.7 & 65.8 \\
    \hline\hline
    IDR~\cite{he2017learning} &95.57&93.89&85.9&64.2&95.4&71.8&58.3\\
    \hline
     W-CNN~\cite{he2018wasserstein} &96.72&95.62&87.0&67.7&96.6&74.9& 62.4\\  
    \hline\hline
        HSST (softmax)& 97.92 & 97.89 & 89.9 & 76.8 & 98.1 & 85.0 & 70.9 \\
    \hline   
        HSST (AM-softmax) & 98.53 & 98.42 & 91.0 & 78.4 & \textbf{98.6} & 85.9 & \textbf{71.7} \\
    \hline   
        HSST (Arc-softmax)& 98.24 & 98.35  & 90.5 & 77.8 & 98.3 & \textbf{86.2} & 71.4 \\  
    \hline 
        HSST (Triplet) & \textbf{98.60} & \textbf{98.58} & \textbf{91.3} & \textbf{83.0} & 98.4  & 85.3 & 70.6 \\
    \hline
    \end{tabular}}
    \end{center}
    \vspace{-2.5em}
\end{table*}

\subsection{Experimental Setting}
We employ two kinds of basic networks, including MobileFaceNet~\cite{chen2018mobilefacenets} in the ablation study and ResNet-50~\cite{wang2017residual} in the cross-dataset experiment. In this paper, all the models are pretrained on the MS1M-v1c dataset. 
For the experiment comparison, we employ the plain training method~\cite{Saxena2016HeterogeneousFR,Liu2016TransferringDR,reale2016seeing} as baseline, which pre-trains the model on MS1M-v1c dataset and fine-tunes it on the NIR-VIS face datasets. The model is trained with two kinds of loss function, including the classification loss (softmax, AM-softmax~\cite{wang2018additive}, and Arc-softmax~\cite{deng2019arcface}) and the feature embedding loss (\textit{i.e.,} triplet~\cite{schroff2015facenet}). Moreover, we also make a comparison with two learning based methods, including IDR~\cite{he2017learning} and W-CNN~\cite{he2018wasserstein}, in the cross-dataset experiment.
In the training stage, the batch size is 64, and the learning rate is initially set as 0.0005 and divided by 10 at 6k and 8k iterations. 
The training is finished at 10k iterations. We employ the size of prototypes as 128, and the weight of moving average as 0.999. 
In the evaluation stage, 512-dimension face representation is extracted from the last fully connected layer of the basic network. The cosine similarity is used as the similarity metric. Note that we only utilize the probe-net for evaluation. 

\begin{figure}[t]
    \centering
    \begin{minipage}[t]{0.47\linewidth}
    \includegraphics[height=3.5cm]{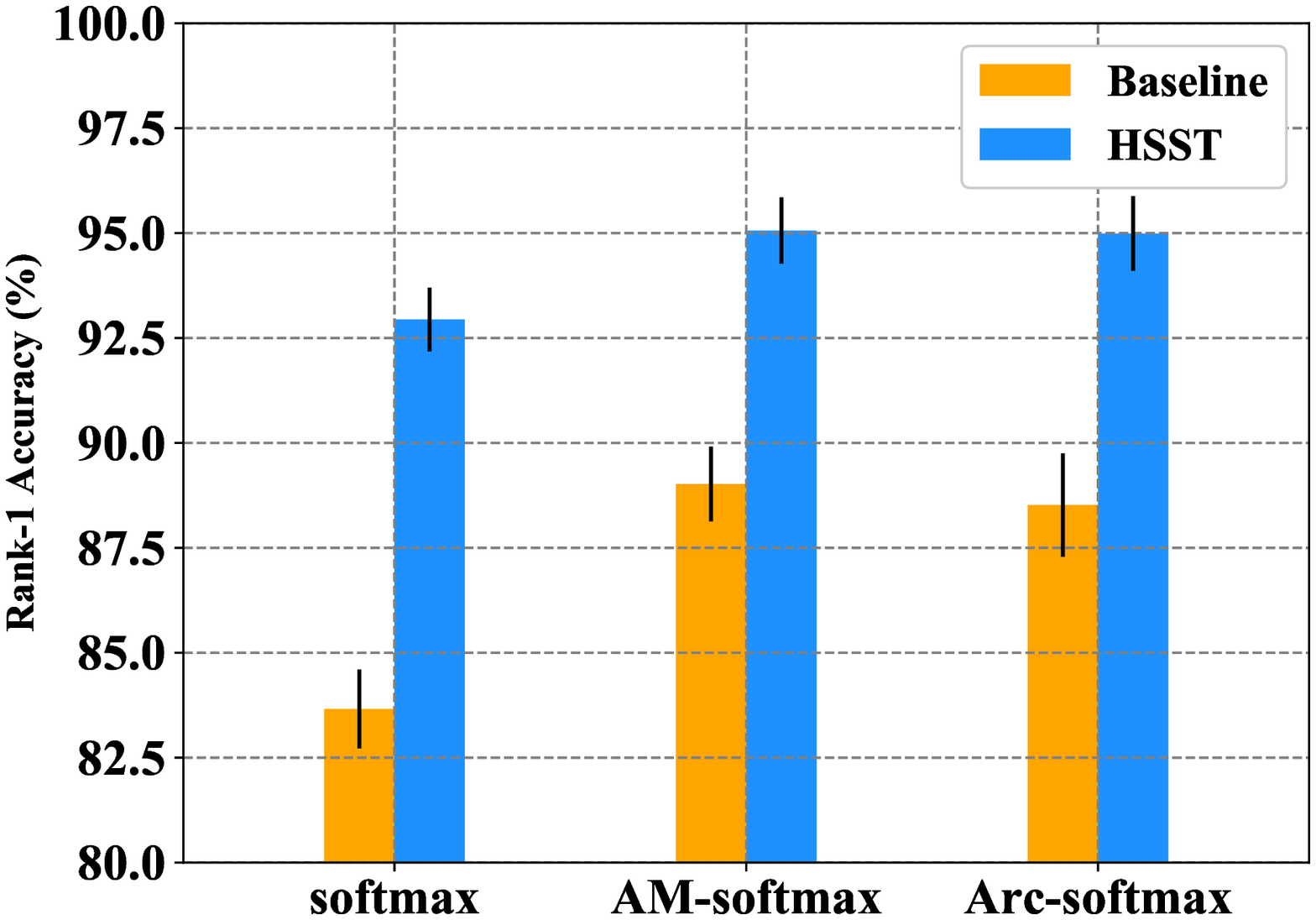}
    \subcaption{Rank-1 Accuracy}
    \label{rank1}
    \end{minipage}
    \begin{minipage}[t]{0.51\linewidth}
    \includegraphics[height=3.5cm]{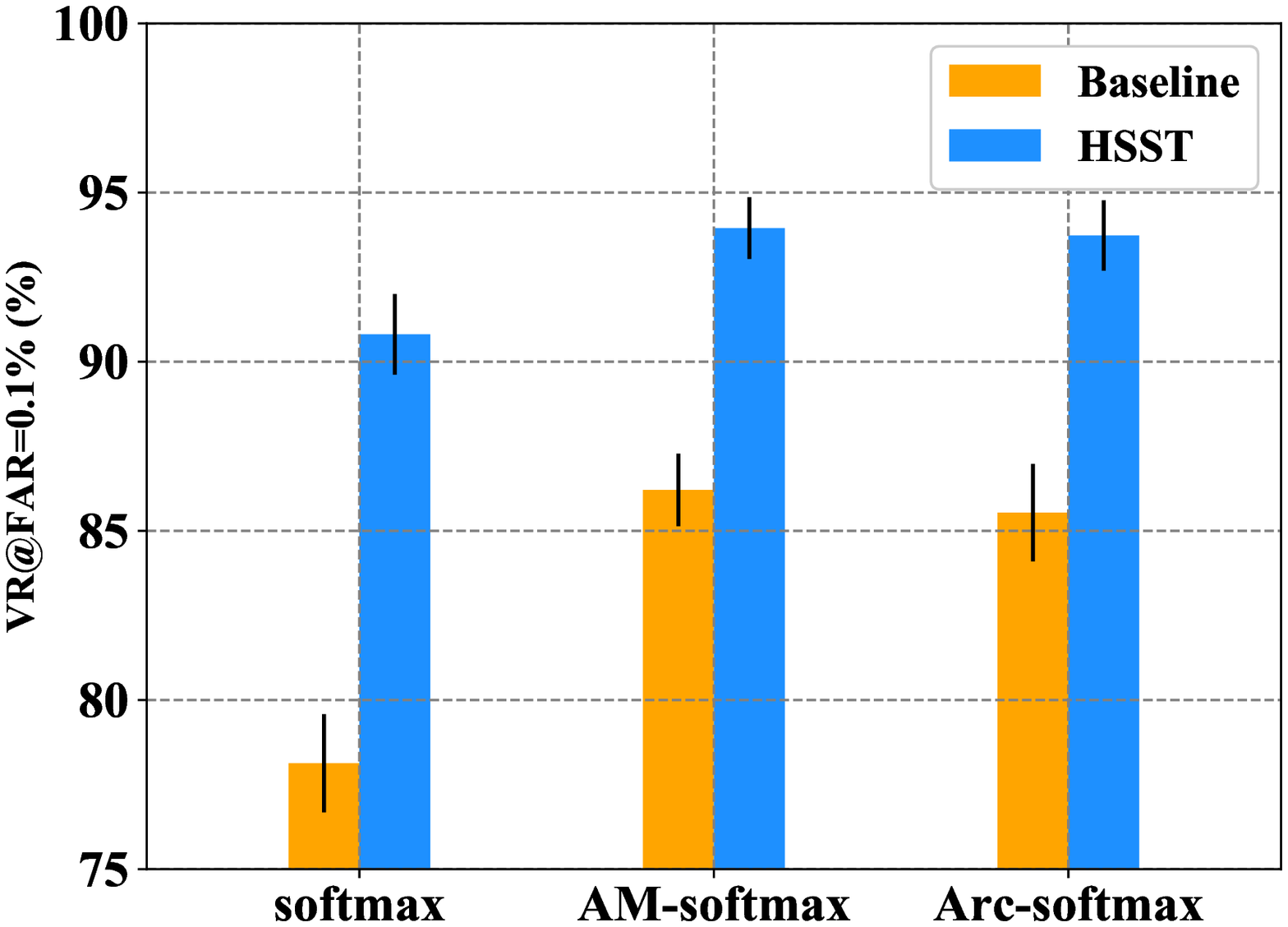}
    \subcaption{VR@FAR=0.1\%}
    \label{flexible_blufr}
    \end{minipage}
    \caption{Performance (\%) comparison on the 10-fold of the synthesized masked CASIA NIR-VIS 2.0 dataset.}
    \label{CASIA_full}
    \vspace{-1em}
\end{figure}

\vspace{-1.5em}
\subsection{Ablation Study}

In ablation study, we use MobileFaceNet~\cite{chen2018mobilefacenets} with softmax loss and train the model on CASIA NIR-VIS 2.0~\cite{li2013casia} dataset. 
We use the standard test protocol in view 2 of CASIA NIR-VIS 2.0 dataset, which contains 10-fold experiments. Each fold contains 357 identities with about 2,500 VIS images and 6,100 NIR images. 
For the evaluation metric, we report the rank-1 accuracy and verification rate at FAR=0.1\%. 
Table~\ref{org_vs_masked} shows the results on the first fold of the non-masked and synthesized masked CASIA NIR-VIS 2.0 dataset. 
From the results of top three rows, we can find the significant performance degradation in NIR-VIS masked face recognition. 
Besides, compared to the bottom three rows, we can conclude that HSST is able to achieve better performance not only on the recognition of the non-masked face but also the masked face. 

In addition, we conduct full 10-fold experiments on CASIA NIR-VIS 2.0 dataset with various training loss functions, and report the mean value and the standard deviation of rank-1 accuracy and verification rate at FAR=1\%. As shown in Fig.~\ref{CASIA_full}, the results show stable improvement brought by HSST, which can validate the superiority of our method on NIR-VIS masked face recognition.

\subsection{Cross-dataset Experiment} 
In this experiment, we adopt ResNet-50 as the backbone and follow the protocol of PCFH~\cite{yu2019pose} to train the model on the first fold of the CASIA NIR-VIS 2.0 dataset. Then, the trained models are evaluated on the Oulu-CASIA NIR-VIS, and the BUAA-VisNir datasets. 

\subsubsection{Results on the 1-fold of CASIA NIR-VIS 2.0 dataset} 
Table~\ref{cross_dataset} shows the performance on the first fold of the synthesized masked CASIA NIR-VIS 2.0 dataset. From the results of~plain training method, we can find triplet loss is able to achieve better performance than softmax loss and its variants. We consider the feature embedding loss is more suitable than the classification loss in such a case of limited training samples. Whatever loss function, a significant performance improvement can be obtained by using HSST.

\subsubsection{Results on Oulu-CASIA NIR-VIS, and BUAA-VisNir datasets}
The Oulu-CASIA NIR-VIS dataset collects images from CASIA and Oulu University, and utilizes all the VIS images from the same identity as the gallery. 
As shown in Table~\ref{cross_dataset}, we only report the verification rate at FAR=1\% and 0.1\%, since all the methods can achieve 100\% rank-1 accuracy. Table~\ref{cross_dataset} also shows the performance comparison on BUAA-VisNir. 
For the evaluation metric, we report the rank-1 accuracy and verification rate at FAR=1\% and 0.1\%. We can observe HSST significantly improve the performance, specifically on the strict false accept rate. Benefit from the design of heterogeneous prototypes and semi-siamese networks, our method can achieve better performance on all the benchmarks. 
The results can demonstrate the better generalization capacity of our method in cross-dataset case.

\vspace{-1em}

\section{Conclusion}
In this paper, we aim to address the NIR-VIS masked face recognition from the perspectives of training data and training method. 
To this end, we propose a heterogeneous semi-siamese training (HSST) to maximize the mutual information shared by the face representation of masked NIR and VIS images from two views, which can facilitate the model to learn domain-invariant face representation that is also robust to the mask occlusion. 
Moreover, we employ a 3D face reconstruction based method to synthesize masked faces for addressing the lack of masked face data. Extensive experiments on three NIR-VIS datasets demonstrate the superiority of our training method over conventional training routine.

\bibliographystyle{IEEEtran}
\bibliography{egbib}

\begin{thebibliography}{10}
\providecommand{\url}[1]{#1}
\csname url@samestyle\endcsname
\providecommand{\newblock}{\relax}
\providecommand{\bibinfo}[2]{#2}
\providecommand{\BIBentrySTDinterwordspacing}{\spaceskip=0pt\relax}
\providecommand{\BIBentryALTinterwordstretchfactor}{4}
\providecommand{\BIBentryALTinterwordspacing}{\spaceskip=\fontdimen2\font plus
\BIBentryALTinterwordstretchfactor\fontdimen3\font minus
  \fontdimen4\font\relax}
\providecommand{\BIBforeignlanguage}[2]{{%
\expandafter\ifx\csname l@#1\endcsname\relax
\typeout{** WARNING: IEEEtran.bst: No hyphenation pattern has been}%
\typeout{** loaded for the language `#1'. Using the pattern for}%
\typeout{** the default language instead.}%
\else
\language=\csname l@#1\endcsname
\fi
#2}}
\providecommand{\BIBdecl}{\relax}
\BIBdecl

\bibitem{Liu2016TransferringDR}
X.~Liu, L.~Song, X.~Wu, and T.~Tan, ``Transferring deep representation for
  nir-vis heterogeneous face recognition,'' \emph{2016 International Conference
  on Biometrics (ICB)}, pp. 1--8, 2016.

\bibitem{he2017learning}
R.~He, X.~Wu, Z.~Sun, and T.~Tan, ``Learning invariant deep representation for
  nir-vis face recognition,'' in \emph{Thirty-First AAAI Conference on
  Artificial Intelligence}, 2017.

\bibitem{lezama2017not}
J.~Lezama, Q.~Qiu, and G.~Sapiro, ``Not afraid of the dark: Nir-vis face
  recognition via cross-spectral hallucination and low-rank embedding,'' in
  \emph{Proceedings of the IEEE conference on computer vision and pattern
  recognition}, 2017, pp. 6628--6637.

\bibitem{yu2019pose}
J.~Yu, J.~Cao, Y.~Li, X.~Jia, and R.~He, ``Pose-preserving cross spectral face
  hallucination.'' in \emph{IJCAI}, 2019, pp. 1018--1024.

\bibitem{duan2020cross}
B.~Duan, C.~Fu, Y.~Li, X.~Song, and R.~He, ``Cross-spectral face hallucination
  via disentangling independent factors,'' in \emph{Proceedings of the IEEE/CVF
  Conference on Computer Vision and Pattern Recognition}, 2020, pp. 7930--7938.

\bibitem{Saxena2016HeterogeneousFR}
S.~Saxena and J.~Verbeek, ``Heterogeneous face recognition with cnns,'' in
  \emph{Proceedings of the European Conference on Computer Vision (ECCV)
  Workshops}, 2016.

\bibitem{reale2016seeing}
C.~Reale, N.~M. Nasrabadi, H.~Kwon, and R.~Chellappa, ``Seeing the forest from
  the trees: A holistic approach to near-infrared heterogeneous face
  recognition,'' in \emph{Proceedings of the IEEE Conference on Computer Vision
  and Pattern Recognition Workshops}, 2016, pp. 54--62.

\bibitem{he2018wasserstein}
R.~He, X.~Wu, Z.~Sun, and T.~Tan, ``Wasserstein cnn: Learning invariant
  features for nir-vis face recognition,'' \emph{IEEE transactions on pattern
  analysis and machine intelligence}, vol.~41, no.~7, pp. 1761--1773, 2018.

\bibitem{juefei2015nir}
F.~Juefei-Xu, D.~K. Pal, and M.~Savvides, ``Nir-vis heterogeneous face
  recognition via cross-spectral joint dictionary learning and
  reconstruction,'' in \emph{Proceedings of the IEEE conference on computer
  vision and pattern recognition workshops}, 2015, pp. 141--150.

\bibitem{song2017adversarial}
L.~Song, M.~Zhang, X.~Wu, and R.~He, ``Adversarial discriminative heterogeneous
  face recognition,'' \emph{arXiv preprint arXiv:1709.03675}, 2017.

\bibitem{Di2018PolarimetricTT}
X.~Di, H.~Zhang, and V.~Patel, ``Polarimetric thermal to visible face
  verification via attribute preserved synthesis,'' \emph{2018 IEEE 9th
  International Conference on Biometrics Theory, Applications and Systems
  (BTAS)}, pp. 1--10, 2018.

\bibitem{Di2019PolarimetricTT}
X.~Di, B.~S. Riggan, S.~Hu, N.~Short, and V.~Patel, ``Polarimetric thermal to
  visible face verification via self-attention guided synthesis,'' \emph{2019
  International Conference on Biometrics (ICB)}, pp. 1--8, 2019.

\bibitem{Di2021MultiScaleTT}
{X. Di, B. S. Riggan, S. Hu, N. Short, and V. Patel}, ``Multi-scale thermal to
  visible face verification via attribute guided synthesis,'' \emph{IEEE
  Transactions on Biometrics, Behavior, and Identity Science}, vol.~3, pp.
  266--280, 2021.

\bibitem{geng2020masked}
M.~Geng, P.~Peng, Y.~Huang, and Y.~Tian, ``Masked face recognition with
  generative data augmentation and domain constrained ranking,'' in
  \emph{Proceedings of the 28th ACM International Conference on Multimedia},
  2020, pp. 2246--2254.

\bibitem{ding2020masked}
F.~Ding, P.~Peng, Y.~Huang, M.~Geng, and Y.~Tian, ``Masked face recognition
  with latent part detection,'' in \emph{Proceedings of the 28th ACM
  International Conference on Multimedia}, 2020, pp. 2281--2289.

\bibitem{du2020semi}
H.~Du, H.~Shi, Y.~Liu, J.~Wang, Z.~Lei, D.~Zeng, and T.~Mei, ``Semi-siamese
  training for shallow face learning,'' in \emph{European Conference on
  Computer Vision}.\hskip 1em plus 0.5em minus 0.4em\relax Springer, 2020, pp.
  36--53.

\bibitem{feng2018joint}
Y.~Feng, F.~Wu, X.~Shao, Y.~Wang, and X.~Zhou, ``Joint 3d face reconstruction
  and dense alignment with position map regression network,'' in
  \emph{Proceedings of the European Conference on Computer Vision (ECCV)},
  2018, pp. 534--551.

\bibitem{li2013casia}
S.~Li, D.~Yi, Z.~Lei, and S.~Liao, ``The casia nir-vis 2.0 face database,'' in
  \emph{Proceedings of the IEEE conference on computer vision and pattern
  recognition workshops}, 2013, pp. 348--353.

\bibitem{chen2009learning}
J.~Chen, D.~Yi, J.~Yang, G.~Zhao, S.~Z. Li, and M.~Pietikainen, ``Learning
  mappings for face synthesis from near infrared to visual light images,'' in
  \emph{2009 IEEE Conference on Computer Vision and Pattern Recognition}.\hskip
  1em plus 0.5em minus 0.4em\relax IEEE, 2009, pp. 156--163.

\bibitem{huang2012buaa}
D.~Huang, J.~Sun, and Y.~Wang, ``The buaa-visnir face database instructions,''
  \emph{School Comput. Sci. Eng., Beihang Univ., Beijing, China, Tech. Rep.
  IRIP-TR-12-FR-001}, 2012.

\bibitem{guo2016ms}
Y.~Guo, L.~Zhang, Y.~Hu, X.~He, and J.~Gao, ``Ms-celeb-1m: A dataset and
  benchmark for large-scale face recognition,'' in \emph{European Conference on
  Computer Vision}.\hskip 1em plus 0.5em minus 0.4em\relax Springer, 2016, pp.
  87--102.

\bibitem{zhang2017faceboxes}
S.~Zhang, X.~Zhu, Z.~Lei, H.~Shi, X.~Wang, and S.~Z. Li, ``Faceboxes: A cpu
  real-time face detector with high accuracy,'' in \emph{2017 IEEE
  International Joint Conference on Biometrics (IJCB)}.\hskip 1em plus 0.5em
  minus 0.4em\relax IEEE, 2017, pp. 1--9.

\bibitem{chen2018mobilefacenets}
S.~Chen, Y.~Liu, X.~Gao, and Z.~Han, ``Mobilefacenets: Efficient cnns for
  accurate real-time face verification on mobile devices,'' in \emph{Chinese
  Conference on Biometric Recognition}.\hskip 1em plus 0.5em minus 0.4em\relax
  Springer, 2018, pp. 428--438.

\bibitem{wang2017residual}
F.~Wang, M.~Jiang, C.~Qian, S.~Yang, C.~Li, H.~Zhang, X.~Wang, and X.~Tang,
  ``Residual attention network for image classification,'' in \emph{Proceedings
  of the IEEE Conference on Computer Vision and Pattern Recognition}, 2017, pp.
  3156--3164.

\bibitem{wang2018additive}
F.~Wang, J.~Cheng, W.~Liu, and H.~Liu, ``Additive margin softmax for face
  verification,'' \emph{IEEE Signal Processing Letters}, vol.~25, no.~7, pp.
  926--930, 2018.

\bibitem{deng2019arcface}
J.~Deng, J.~Guo, N.~Xue, and S.~Zafeiriou, ``Arcface: Additive angular margin
  loss for deep face recognition,'' in \emph{Proceedings of the IEEE Conference
  on Computer Vision and Pattern Recognition (CVPR)}, 2019, pp. 4690--4699.

\bibitem{schroff2015facenet}
F.~Schroff, D.~Kalenichenko, and J.~Philbin, ``Facenet: A unified embedding for
  face recognition and clustering,'' in \emph{Proceedings of the IEEE
  conference on computer vision and pattern recognition}, 2015, pp. 815--823.

\end{thebibliography}

\end{document}